\documentclass[sigconf,9pt,nonacm]{acmart}
\setcopyright{none}
\renewcommand\footnotetextcopyrightpermission[1]{}
\usepackage{comment}
\usepackage{graphicx}
\usepackage{subcaption}
\usepackage{enumitem}
\usepackage{xcolor}    
\usepackage{soul}      

\begin{document}

\title{Distributed Cross-Channel Hierarchical \\ Aggregation for Foundation Models}


\author{Aristeidis Tsaris}
\authornote{Corresponding author: tsarisa@ornl.gov}
\affiliation{%
  \institution{Oak Ridge National Laboratory}
  \city{Oak Ridge}
  \state{Tennessee}
  \country{USA}
}

\author{Isaac Lyngaas}
\affiliation{%
  \institution{Oak Ridge National Laboratory}
  \city{Oak Ridge}
  \state{Tennessee}
  \country{USA}
}

\author{John Lagregren}
\affiliation{%
  \institution{Oak Ridge National Laboratory}
  \city{Oak Ridge}
  \state{Tennessee}
  \country{USA}
}

\author{Mohamed Wahib}
\affiliation{%
  \institution{RIKEN Center for Computational Science}
  \city{Kobe}
  \country{Japan}
}

\author{Larry York}
\affiliation{%
  \institution{Oak Ridge National Laboratory}
  \city{Oak Ridge}
  \state{Tennessee}
  \country{USA}
}

\author{Prasanna Balaprakash}
\affiliation{%
  \institution{Oak Ridge National Laboratory}
  \city{Oak Ridge}
  \state{Tennessee}
  \country{USA}
}

\author{Dan Lu}
\affiliation{%
  \institution{Oak Ridge National Laboratory}
  \city{Oak Ridge}
  \state{Tennessee}
  \country{USA}
}

\author{Feiyi Wang}
\affiliation{%
  \institution{Oak Ridge National Laboratory}
  \city{Oak Ridge}
  \state{Tennessee}
  \country{USA}
}

\author{Xiao Wang}
\affiliation{%
  \institution{Oak Ridge National Laboratory}
  \city{Oak Ridge}
  \state{Tennessee}
  \country{USA}
}

\begin{abstract}

Vision-based scientific foundation models hold significant promise for advancing scientific discovery and innovation. This potential stems from their ability to aggregate images from diverse sources—such as varying physical groundings or data acquisition systems—and to learn spatio-temporal correlations using transformer architectures. However, tokenizing and aggregating images can be compute-intensive, a challenge not fully addressed by current distributed methods. In this work, we introduce the Distributed Cross-Channel Hierarchical Aggregation (D-CHAG) approach designed for datasets with a large number of channels across image modalities. Our method is compatible with any model-parallel strategy and any type of vision transformer architecture, significantly improving computational efficiency. We evaluated D-CHAG on hyperspectral imaging and weather forecasting tasks. When integrated with tensor parallelism and model sharding, our approach achieved up to a 75\% reduction in memory usage and more than doubled sustained throughput on up to 1,024 AMD GPUs on the Frontier Supercomputer.

\end{abstract}

\keywords{Computing methodologies; Machine learning algorithms; Parallel algorithms; Distributed deep learning}

\maketitle

\section{Introduction} \label{sec:intro}

Vision Transformers (ViTs) have been widely adopted in multi-modal foundation models (FMs), particularly those combining language and image modalities. More recently, ViTs have been applied to scientific imagery applications, including research in Earth system science and geospatial analysis \cite{chen2022scaling, xiong2024all, cha2023billionscale}.

Despite their success, scientific FMs have not yet scaled to the size of large language models (LLMs). While the largest LLMs are trained on over one trillion parameters \cite{10.5555/3586589.3586709}, the largest ViT for natural images currently reaches only 22 billion parameters \cite{dehghani2023scaling}. Most vision FMs focus on natural images, but training scientific FMs presents different challenges—chiefly, the complexity of the data. This added complexity makes scientific FMs especially well-suited for benefits from large-scale modeling. A notable example is ORBIT \cite{wang2024orbitoakridgebase}, a large scientific FM scaled to 113 billion parameters.

One major factor contributing to the complexity of scientific data is the number of channels. In many-body scientific simulations, outputs often consist of large sets of independent 2D images, which are encoded as separate channels in foundation models \cite{nguyen2023climax, bodnar2024foundationmodelearth, nguyen2024scalingtransformerneuralnetworks, wang2024orbitoakridgebase}. For instance, climate simulations \cite{gmd-9-1937-2016} and weather observational data \cite{https://doi.org/10.1002/qj.3803} may have over 100 independent channels. In E3SM biogeochemistry simulations, outputs can reach over 500 channels \cite{YUAN2023102145}. Similarly, hyperspectral images—which contain hundreds of contiguous spectral bands—can result in more than 500 channels \cite{doi:10.1126/science.abe0722}. Such high-dimensional data appears across numerous scientific domains, including geospatial analysis, medicine, and biology. 

The state-of-the-art method for reducing the quadratic complexity associated with the number of channels in the self-attention mechanism—the ViT component of a foundation model (FM)—is to introduce cross-attention between channels \cite{bodnar2024foundationmodelearth}. The cross-attention layer enables effective learning along the channel dimension by reducing or aggregating the channels into a single representation before passing the data to the self-attention layer, which operates in the spatial dimension.

To scale FMs for scientific imaging, several distributed methods have been employed. For instance, data parallelism (DP) scales with the dataset size, tensor parallelism (TP) scales the embedding dimension of the model, and sequence parallelism (SP) scales the sequence length of the tokens. However, none of these methods address scaling along the channel dimension—this is the specific challenge our approach targets.

We propose the Distributed Cross-Channel Hierarchical Aggregation (D-CHAG) method for foundation models, which distributes tokenization and implements a hierarchical strategy for channel aggregation. Our approach reduces memory usage by up to 70\%, enabling the execution of extremely large models on multi-channel datasets—capabilities not possible with existing methods. D-CHAG is designed to be general and can be integrated with other parallelism techniques. In this work, we combine D-CHAG with TP, with the latter serving as our baseline. We evaluated D-CHAG on two applications using distinct architectures: (1) weather forecasting with an image-to-image translation model, and (2) plant phenotype analysis using a self-supervised masked autoencoder on hyperspectral images. When scaled to 1,024 GPUs with real hyperspectral data, our method achieved up to a 239\% increase in TFLOPs/sec compared to TP alone.
It is worth noting that our method is general enough to be combined with any ViT architecture, and any of the current model parallel methods for transformer, such as TP and SP. Additionally, our findings can be expand on beyond single multi-channel datasets, as the same aggregation scheme has be used in FMs to fuse across different modalities \cite{xu2023multimodallearningtransformerssurvey}.

Our contributions are the following:
\begin{itemize}[leftmargin=2.5mm]
    \item We identify limitations in existing distributed methods when scaling foundation models (FMs) to scientific multi-channel datasets.
    \item We propose a novel \textbf{Distributed Cross-Channel Hierarchical Aggregation (D-CHAG)} method that reduces redundant computation and improves memory efficiency by distributing the tokenization and channel aggregation stages. Our hierarchical design minimizes communication during the forward pass and requires no communication in the backward pass. D-CHAG is compatible with TP, SP, and any ViT architecture.
    \item D-CHAG achieves up to a \textbf{70\% reduction in memory usage} compared to TP alone, enabling more efficient training of large-scale models. 
    \item We demonstrate that D-CHAG allows scaling to larger model sizes on multi-channel datasets. While TP scales the embedding dimension, it alone cannot handle the increased overhead from tokenization and channel aggregation. D-CHAG removes this bottleneck.
    \item By combining D-CHAG with TP, FSDP, and DP, we achieve over \textbf{2× sustained throughput} compared to TP alone, scaling up to \textbf{1,024 AMD GPUs} on the Frontier Supercomputer using real hyperspectral image datasets.
    \item We validate D-CHAG on two scientific workloads—weather forecasting and masked prediction on plant hyperspectral images—with less than 1\% degradation in solution quality. 
\end{itemize}

\section{Background and Motivation} \label{sec:bkg}

There has been significant recent effort to develop foundation models (FMs) for a variety of scientific domains. Due to the inherent complexity of scientific data, various fusion techniques are employed to jointly learn spatio-temporal representations. Additionally, several distributed training approaches originally developed for large language models (LLMs) have been adapted for this purpose.

\subsection{The Channel Aggregation Module}

One of the most impactful applications of scientific foundation models (FMs) has been in the domain of weather forecasting \cite{nguyen2023climax, bodnar2024foundationmodelearth, nguyen2024scalingtransformerneuralnetworks, wang2024orbitoakridgebase}. These architectures typically employ a cross-attention module to aggregate information across the channel dimension before passing it to the Vision Transformer (ViT) component, which uses self-attention to model spatial relationships.

\begin{figure}[h!]
\centering
\hspace*{0cm}\includegraphics[width=0.85\linewidth]{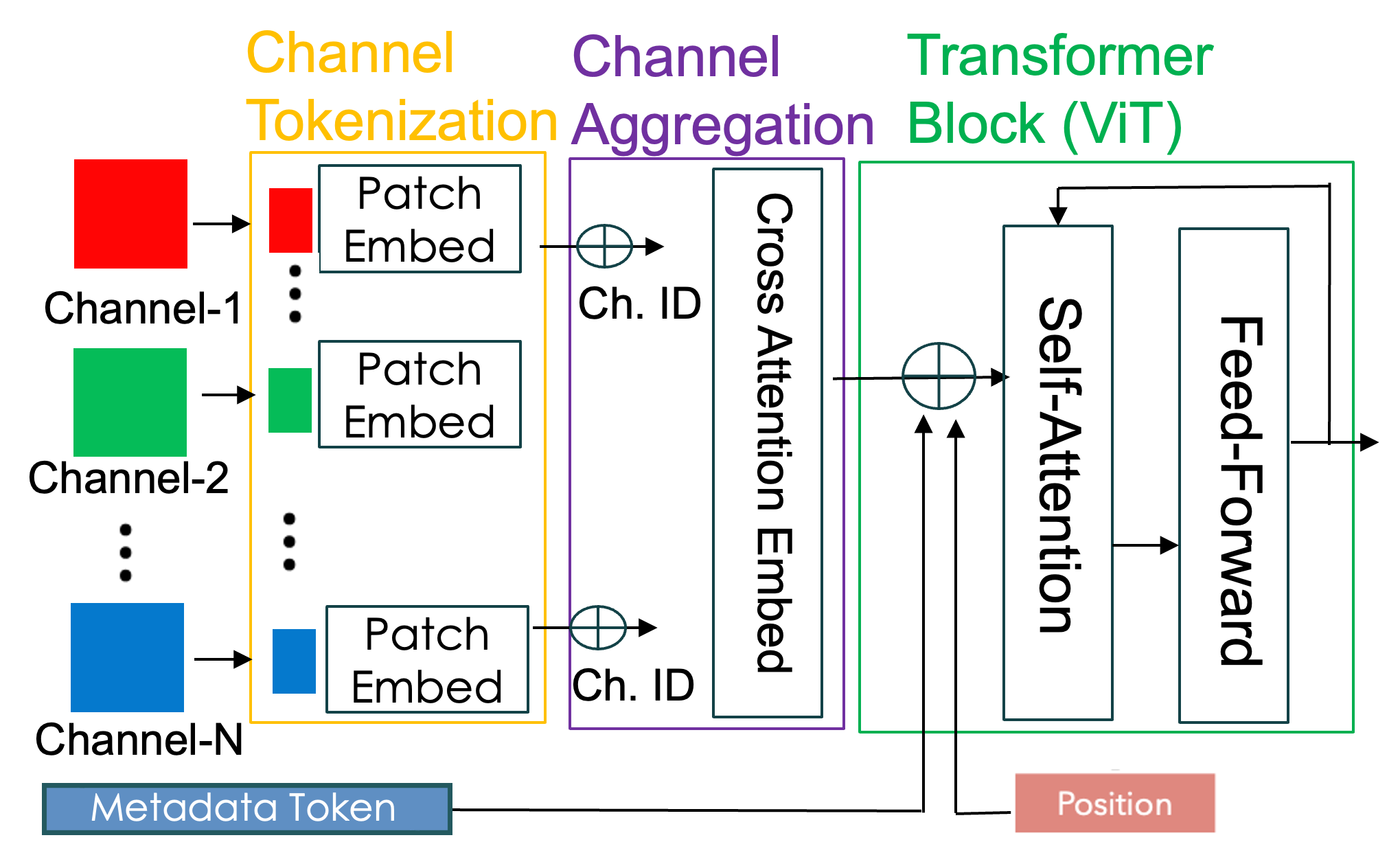}
\caption{The diagram shows the generic model architecture used in this work.}
\label{fig:diagram-arch}
\end{figure}

A simplified version of these architectures is illustrated in Figure~\ref{fig:diagram-arch}, which also forms the basis of the model used in this work for development and performance analysis. The inputs are 2D images with many channels—such as different physical variables in weather forecasting or different wavelengths in hyperspectral imaging. These images are divided into patches, and 2D convolution layers are applied to each patch for tokenization. Each 2D patch from each channel is tokenized independently.

The resulting tokens are then passed to a channel aggregation module. In our architecture, a single cross-attention layer aggregates the information across channels, reducing them to a single representation. This aggregated representation is then passed to the Vision Transformer (ViT) component, where dense self-attention layers are used in the transformer blocks.

Special tokens are incorporated at various stages of the model. For example, during channel aggregation, each channel is assigned a special ID token, which can represent channels from the same or different modalities. A positional token encodes the spatial location of each patch in the original image. These are concatenated with a metadata token—typically representing contextual information like time and geospatial location in weather forecasting—before being passed to the ViT. This design allows the model to learn collectively from all input channels.

There are several reasons why scientific foundation models in climate and related domains use cross-attention for channel fusion instead of feeding all channels directly into a ViT. Self-attention has quadratic memory complexity with respect to sequence length. Aggregating the channels before the ViT moves this complexity from the self-attention layer (which operates on spatial tokens) to the cross-attention layer (which operates on channels). For datasets with a large number of channels, this step is essential for fitting the model in memory, as self-attention already incurs high memory costs in the spatial dimension.

From a modeling perspective, the channel aggregation module also improves learning from diverse physical inputs or heterogeneous data sources, such as variables recorded at different resolutions or with different channel counts. It further allows the model to generalize or fine-tune on subsets of the original channel dimensions while still leveraging the full model capacity, greatly enhancing its flexibility for real-world deployment.

\subsection{Distributed Methods for FMs}

Several methods have been developed to scale foundation models (FMs) across a large number of GPUs, though most of these methods were not specifically designed for vision FMs and are instead applicable to any transformer architecture.
Data-parallel (DP) is the most commonly used distributed method in deep learning, employed to distribute the data size. Each DP worker holds a copy of the entire model but processes different portions of the data. During the forward pass, there is no communication, while lightweight communication via AllReduce occurs at the end of the backward pass. DP scales efficiently because computation grows with communication.

The main limitation of the DP method is that model size is constrained by available GPU memory, as each GPU holds a copy of the entire model. The model-sharding method, such as FSDP \cite{zhao2023pytorchfsdpexperiencesscaling}, addresses this limitation by distributing the gradients, model parameters, and optimizer states across GPUs. It replaces AllReduce communication with more frequent AllGather and ReduceScatter operations. Model-sharding methods are not strictly considered model-parallel, as their primary goal is still to scale the data size. However, by removing much of the redundant computation present in DP, they enable the training of larger models that would otherwise exceed the memory limits of a single GPU. Despite this, model-sharding methods also face their own size limitations, as demonstrated by \cite{wang2024orbitoakridgebase}, because at some point, the entire model parameters must fit into the memory of a single GPU.

Model-parallel methods are usually less architecture-agnostic compared to model-sharding and data-parallel methods. For example, sequence-parallelism (SP) is particularly relevant to transformer architectures, where its primary function is to distribute the sequence length of the tokens. On the other hand, tensor-parallelism (TP) has been essential in scaling the model size of FMs, and has been extensively used in large language models (LLMs) and Vision Transformers (ViTs). TP distributes the embedding dimension to scale the model size by increasing the number of parameters in the transformer block.

As mentioned earlier, scientific images often contain a large number of channels, and tokenization and aggregation can become significant bottlenecks when scaling FMs trained on these images. The D-CHAG method addresses this issue and is complementary to TP and SP methods.

\section{The Distributed Cross-Channel Hierarchical Aggregation Method} \label{sec:method} 

In this section, we first present two separate approaches for reducing memory usage in foundation models (FMs) with a large number of channels. We then combine these approaches to form the D-CHAG method and demonstrate how it can be integrated with other distributed strategies for FMs.

\subsection{Distributed Channel Tokenization} \label{dist-token-method}

\begin{figure}[h!]
\centering
\hspace*{0cm}\includegraphics[width=0.85\linewidth]{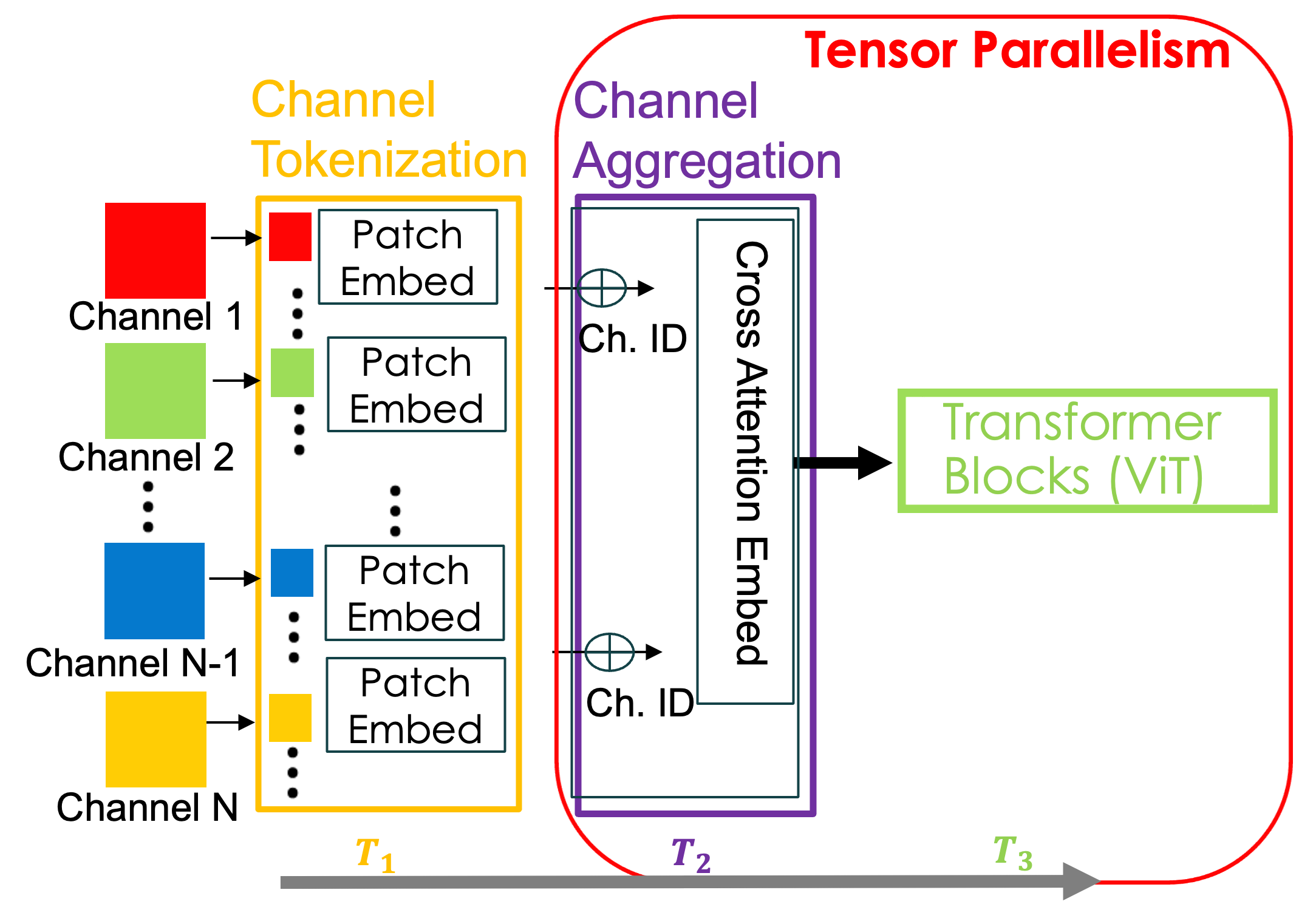}
\hspace*{0cm}\includegraphics[width=0.85\linewidth]{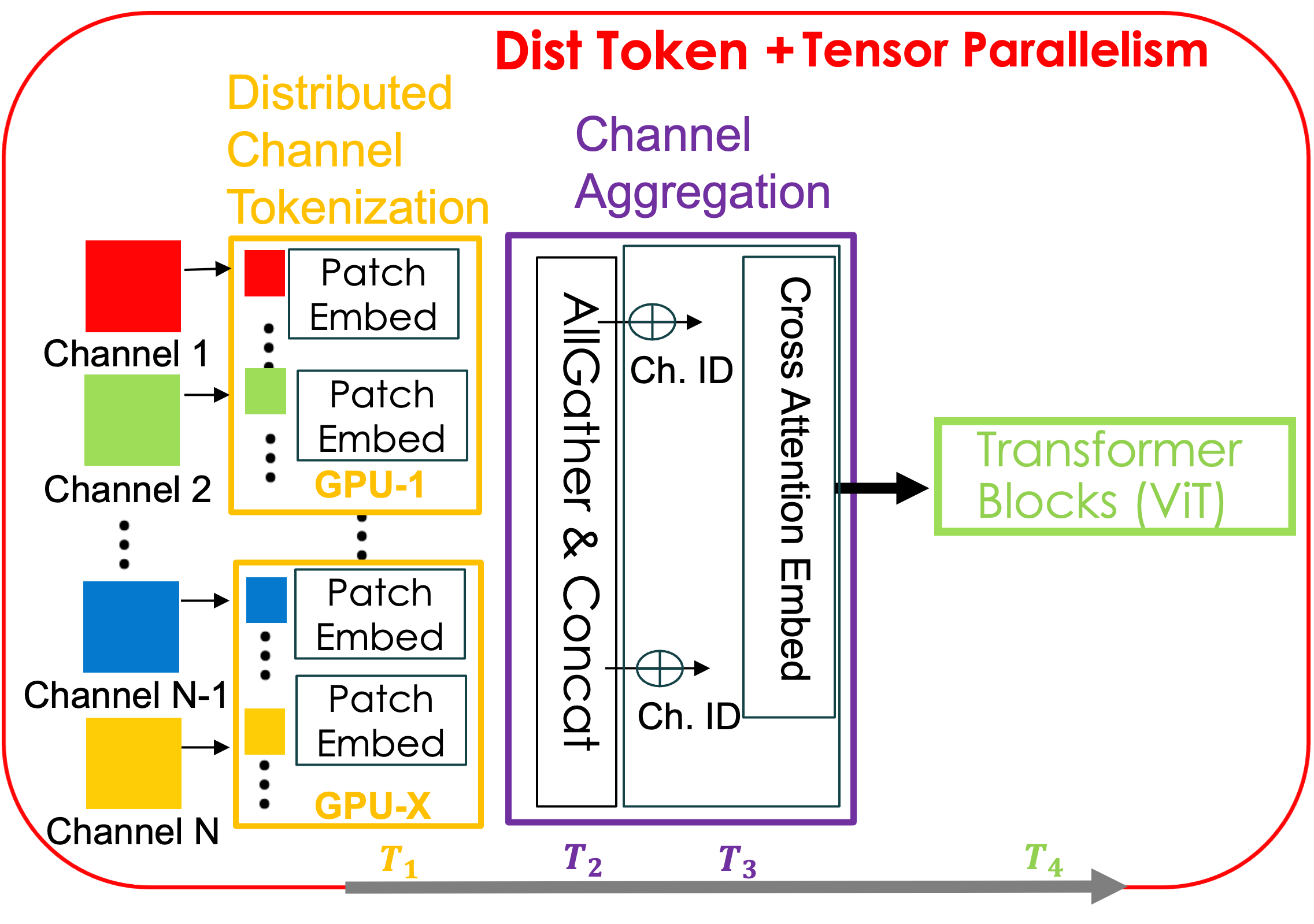}
\caption{The top diagram shows a schematic picture for TP applied on the FM architecture used for this work. The bottom diagram illustrates a distributed tokekization approach.}
\label{fig:diagram-arch-tp}
\end{figure}

The top plot of Figure~\ref{fig:diagram-arch-tp} illustrates the application of tensor parallelism (TP) to the foundation model (FM). Both the Vision Transformer (ViT) and the channel aggregation module consist of attention mechanisms, and since TP scales the embedding space of attention, we apply TP to both components in this work. As shown in the top diagram, channel tokenization is redundantly computed across all TP ranks.

The simplest approach is to distribute tokenization across the TP ranks, as illustrated in Figure~\ref{fig:diagram-arch-tp}. During the forward pass, each TP rank tokenizes only a subset of the input channels. An \emph{AllGather} operation is then required before the channel aggregation step to enable cross-attention across all channels. This approach should, in principle, reduce the tokenization computation cost per GPU, although it introduces additional overhead due to the \emph{AllGather} operation across both the channel and spatial dimensions.

\subsection{Hierarchical Cross-Channel Aggregation}

\begin{figure}[h!]
\centering
\hspace*{0cm}\includegraphics[width=0.85\linewidth]{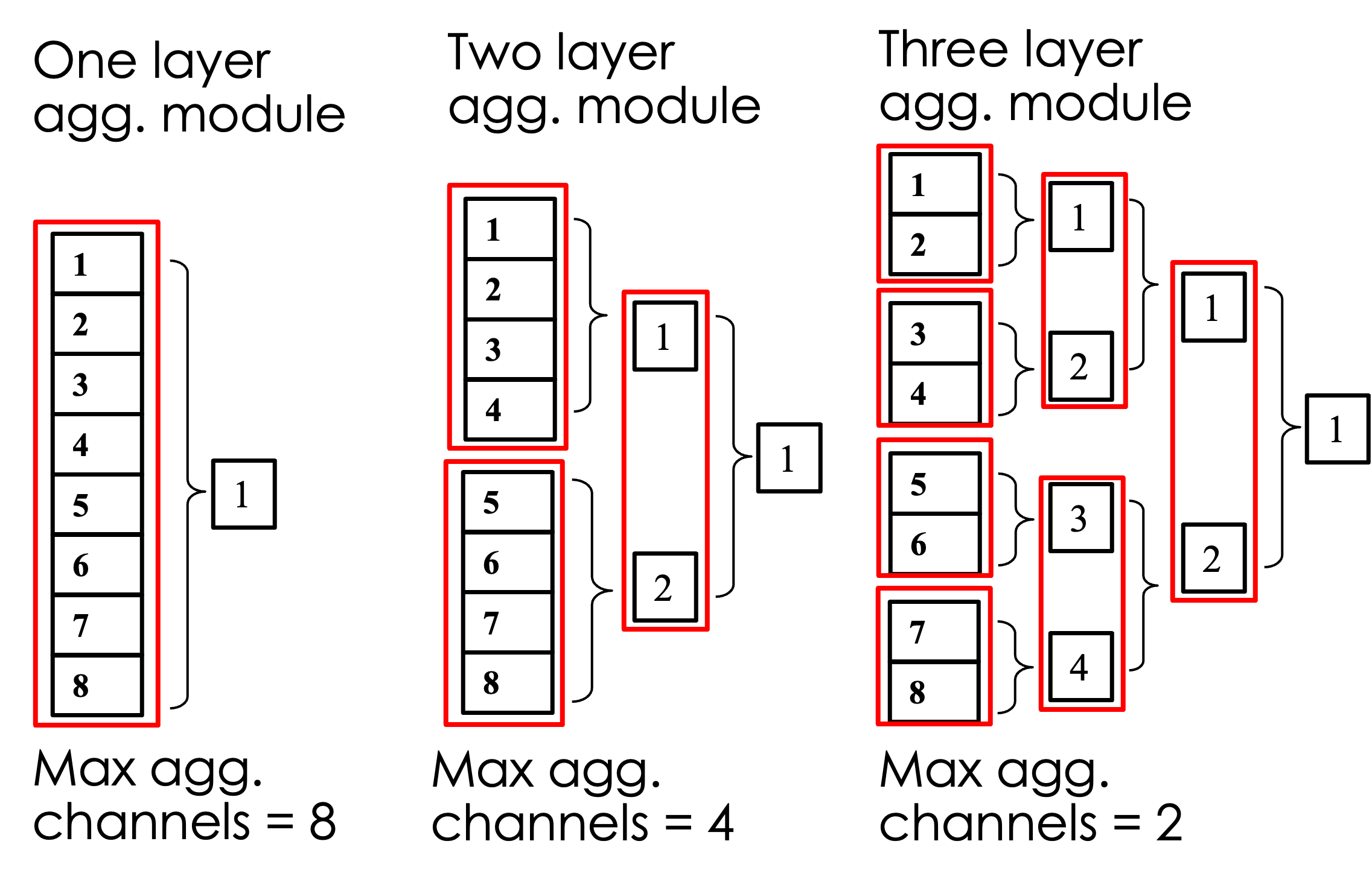}
\caption{Sketch of hierarchical channel aggregation for data with eight channels.}
\label{fig:diagram-hier}
\end{figure}

Another major contributor to the model’s memory footprint—aside from the ViT module—is the channel aggregation module. The memory usage of cross-attention scales quadratically with the number of channels. As mentioned earlier, self-attention in ViTs exhibits a similar quadratic dependency, but several approximations, such as sparse attention, can be applied to mitigate this. However, these sparse attention techniques are not directly applicable to cross-attention due to the uneven nature of the input and output variables. Moreover, tensor parallelism (TP) does not address the channel-wise complexity of cross-attention, as it distributes computation across the embedding dimension rather than the channel dimension.

One way to mitigate this limitation is by adding multiple cross-attention layers to the channel aggregation module and performing aggregation hierarchically. This strategy can reduce the complexity from quadratic to linear with respect to the number of channels.

Figure~\ref{fig:diagram-hier} illustrates three configurations of the channel aggregation module using eight channels, with up to three layers of cross-attention. The first configuration represents the baseline, consisting of a single cross-attention layer processing all eight channels at once. The second configuration employs a two-layer hierarchy, where each cross-attention layer handles up to four channels. In the third example, a three-layer hierarchy reduces the maximum number of channels per layer to just two.

The main advantage of this approach is a reduced memory footprint per cross-attention layer, since each layer processes fewer channels. However, increasing the number of layers does lead to a larger overall model size and higher memory usage. This trade-off becomes more favorable for datasets with a large number of channels, where the quadratic memory cost of standard cross-attention is larger. Additionally, the parameter overhead introduced by the extra layers can be minimized by replacing intermediate cross-attention layers with lightweight linear layers, while keeping only the final layer as full cross-attention.

\subsection{Combine Hierarchical Aggregation \\with Distributed Tokenization} \label{dchag-method}

\begin{figure}[h!]
\centering
\hspace*{0cm}\includegraphics[width=1.0\linewidth]{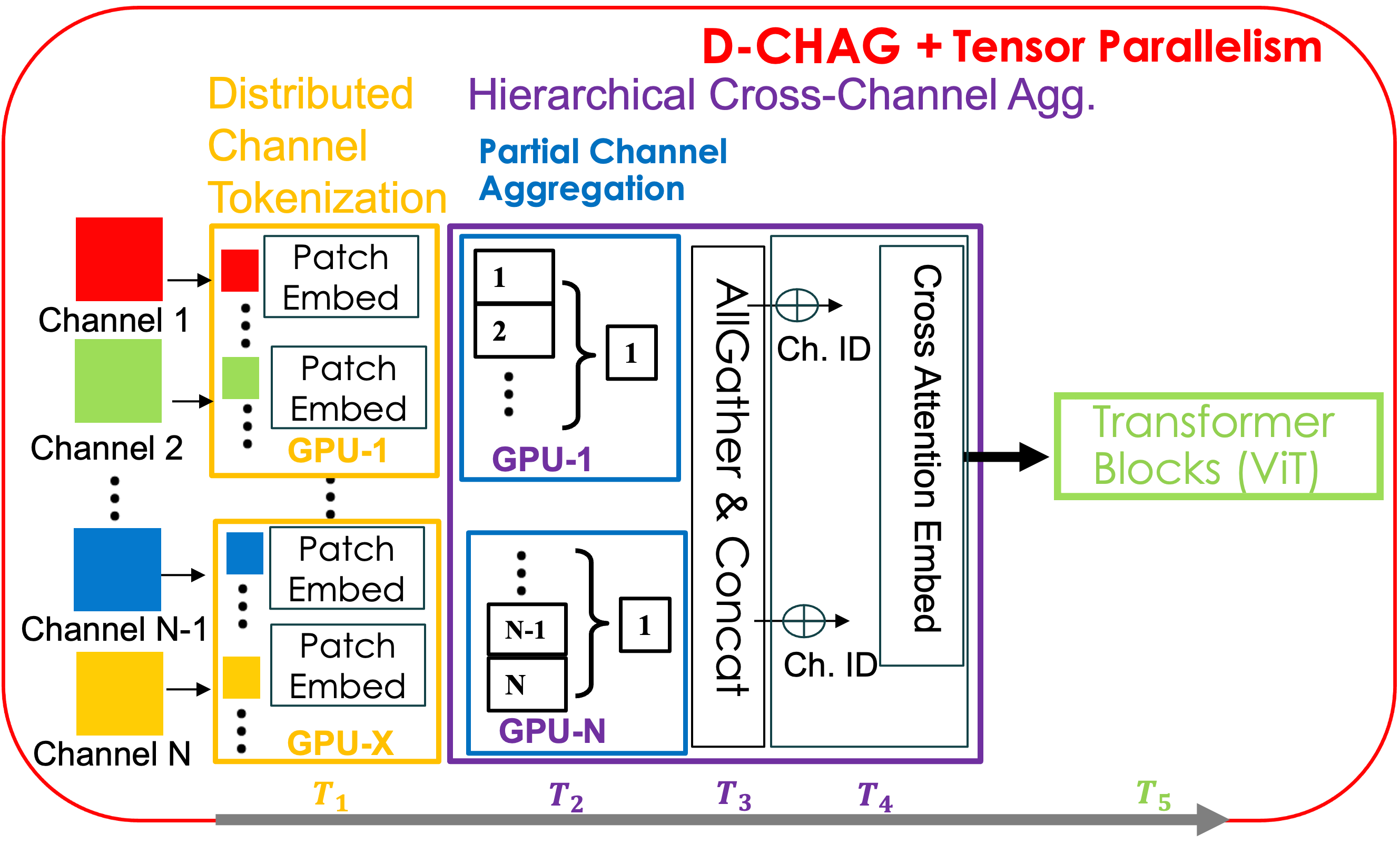}
\caption{Sketch of the D-CHAG method applied in the base architecture.}
\label{fig:diagram-arch-dchavit}
\end{figure}

So far, we've discussed two approaches that, while offering advantages, also have some drawbacks. Specifically, the distributed tokenization approach suffers from high communication overhead between TP ranks and does not address the issue of large memory footprints in the channel dimension. Meanwhile, the hierarchical cross-channel aggregation approach increases the number of model parameters per GPU. The D-CHAG method combines both approaches in a distributed fashion.

Figure~\ref{fig:diagram-arch-dchavit} illustrates the D-CHAG approach. Each TP rank tokenizes 2D images from a subset of the total channels. Since each GPU holds only a fraction of the full channel set, we perform channel aggregation locally on those channels—this module is referred to as the partial-channel aggregation module. After aggregating channels within each TP rank, we gather the outputs and perform a final aggregation using cross-attention. This requires only an \emph{AllGather} operation during the forward pass; during the backward pass, we gather only the relevant gradients for each GPU, avoiding any additional communication.

With the D-CHAG approach, we leverage the strengths of both distributed tokenization and hierarchical channel aggregation while mitigating their weaknesses. By distributing the hierarchical channel aggregation across TP ranks, we reduce the \emph{AllGather} communication to a single channel per TP rank, with no communication needed during the backward pass. Furthermore, we preserve the benefits of reducing the number of channels per aggregation layer by adding depth to the model, while distributing the extra model parameters across TP ranks via the partial-channel aggregation modules.

It is also worth noting that the partial-channel aggregation modules offer several tunable parameters. For example, the depth of the hierarchical approach can be adjusted based on performance gains. Additionally, the cross-attention layers can be replaced with linear layers. As long as the final channel aggregation layer—shared across all GPUs—remains a cross-attention layer, we do not expect any degradation in performance. In all experiments presented in this work, the final layer of the channel aggregation module has been implemented as a cross-attention layer. We refer to variants using linear layers in the partial aggregation module as \emph{D-CHAG-L}, and those using cross-attention as \emph{D-CHAG-C}.

Finally, D-CHAG is fully integrated with TP. The final cross-attention layer is shared across all TP ranks, and because the data are fully synchronized across GPUs at that point, we can distribute the embedding space similarly to how we distribute it in the downstream transformer block modules.

\subsection{Hybrid Parallelism for D-CHAG}

\begin{figure}[h!]
\centering
\hspace*{0cm}\includegraphics[width=1.0\linewidth]{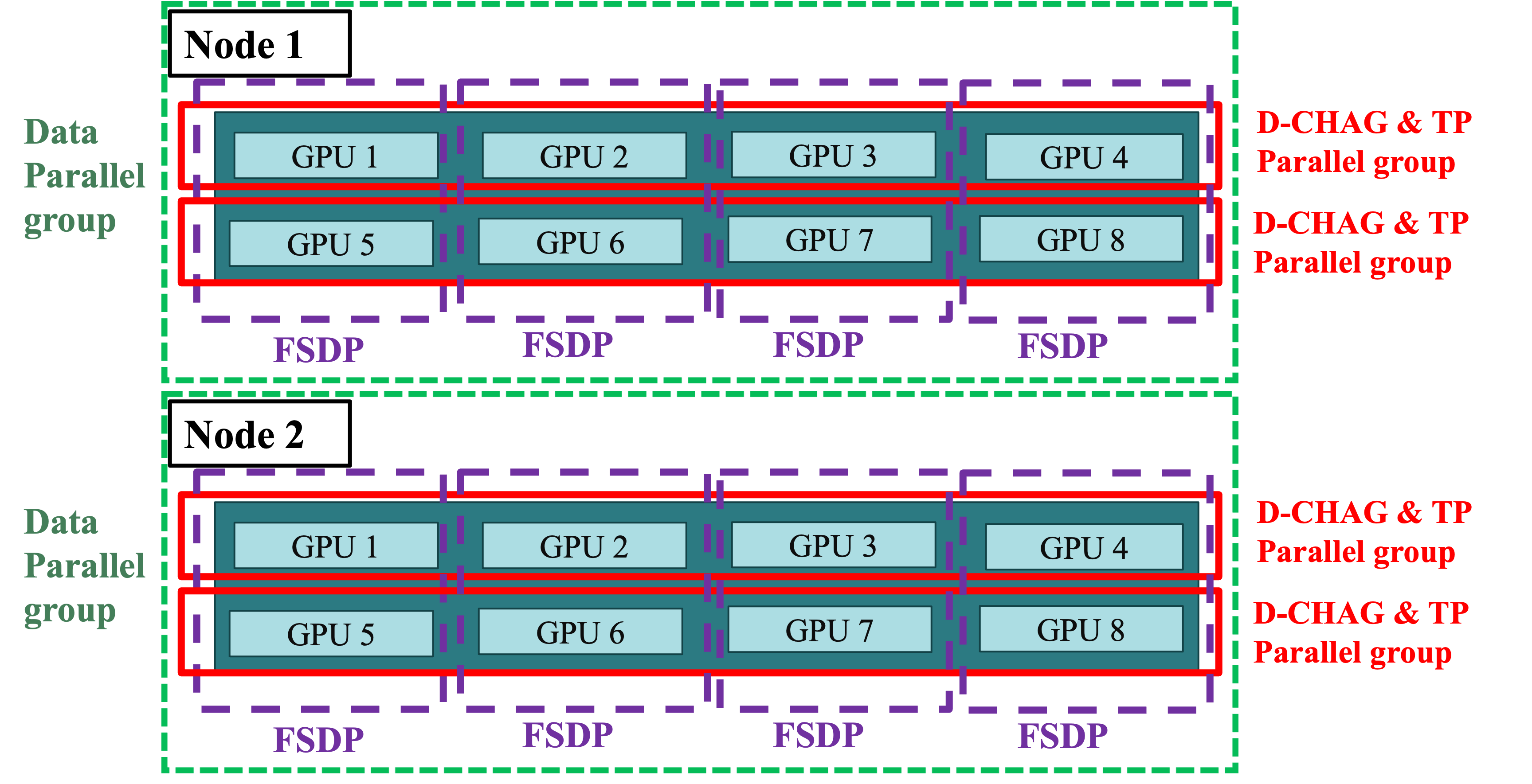}
\caption{Sketch of Hybrid D-CHAG, combining model-sharding and data-parallel.}
\label{fig:diagram-scale}
\end{figure}

We can also integrate the D-CHAG and TP methods with model-sharding and data-parallel techniques. In both cases, we used native PyTorch implementations—namely, Fully Sharded Data Parallel (FSDP) and Data Parallel (DP). Figure~\ref{fig:diagram-scale} illustrates how FSDP and DP can be integrated with D-CHAG and TP. Since D-CHAG already operates over TP ranks, the D-CHAG and TP groups are identical. FSDP, layered on top of TP, helps further reduce the memory footprint by sharding model parameters, gradients, and optimizer states.

As we've seen, TP distributes specific parts of the model—such as the embedding space—across devices while processing the same batches of data. FSDP, by contrast, is more architecture-agnostic and operates over different batches, offering memory savings across a wider portion of the model. As expected, FSDP incurs less communication overhead than TP, while DP has the least communication cost but provides no memory reduction. These distributed strategies belong to different parallel groups, as illustrated in Figure~\ref{fig:diagram-scale}.

\subsection{Method Versatility}

The D-CHAG method is generalizable to any FM architecture that combines fusion through cross-attention with ViT. In fact, using cross-attention to fuse data across different modalities—while relying on self-attention to capture spatial correlations within the same modality—is a common strategy \cite{xu2023multimodallearningtransformerssurvey}. D-CHAG can also be applied to any self-supervised method, as it only modifies the input to the ViT module, without altering the decoder modules.

Moreover, the D-CHAG method is agnostic to the specific ViT architecture and, to some extent, to the fusion module as well. For instance, Aurora \cite{bodnar2024foundationmodelearth}, one of the latest and most advanced FMs for weather prediction, employs the Perceiver architecture \cite{jaegle2021perceivergeneralperceptioniterative} as the fusion module and a Swin Transformer \cite{liu2021swintransformerhierarchicalvision} in place of a standard ViT. The Perceiver, being a more computationally intensive cross-attention-based module, is likely to show even greater performance benefits from D-CHAG compared to the simpler single cross-attention layer we benchmarked in this work.
In addition, the Swin Transformer applies a hierarchical approach to self-attention, enabling it to handle longer sequence-length tokens. This increases the workload for tokenization and channel aggregation, suggesting that D-CHAG could yield even greater benefits in such settings, especially when approximations are applied to the self-attention layers.

Finally, while we have integrated D-CHAG with TP in this work, the method could also be used with other model-parallel strategies. For example, Sequence Parallelism (SP) could operate on the same model segments—just before the self-attention layers—to distribute sequence length. In that scenario, although performance characteristics might differ, our method would remain directly applicable, enabling tokenization and hierarchical aggregation to be distributed along the axis in which the data are fused.

\section{Performance Analysis}

In this section, we present a performance analysis across three settings: on a single GPU, on multiple GPUs using tensor parallelism (TP), and on multiple GPUs using TP combined with the D-CHAG method.

\subsection{The Frontier Architecture}
All the experiments were performed using the Frontier Supercomputer \cite{Frontier} at the Oak Ridge Leadership Computing Facility. Each Frontier node has a single 64-core AMD EPYC CPU and four AMD Instinct MI250X GPU accelerators. The MI250X GPU is comprised of two Graphics Compute Dies (GCDs), connected with Infinity Fabric CPU-GPU, while the four MI250X GPUs are connected with Infinity Fabric GPU-GPU of 50GB/s. The system identifies each GCD independently, so from the application perspective, it can be considered that each node has 8 GPUs, each with 64 GB of high-bandwidth memory. For simplicity, we will use the term GPU when referring to a GCD. The nodes are connected via a Slingshot-11 interconnect with 100GB/s to a total of 9408 nodes, making it the first true exascale machine. For the software stack, we used Pytorch 2.4, with Flash-Attention-v2 (FA2) \cite{dao2023flashattention2} on all experiments, ROCm v5.7.0, MIOpen v2.19.0, RCCL v2.13.4 with libfabric v1.15.2 plugin.

\subsection{Single GPU Performance Analysis}

\begin{figure}[h!]
\centering
\hspace*{0cm}\includegraphics[width=1.0\linewidth]{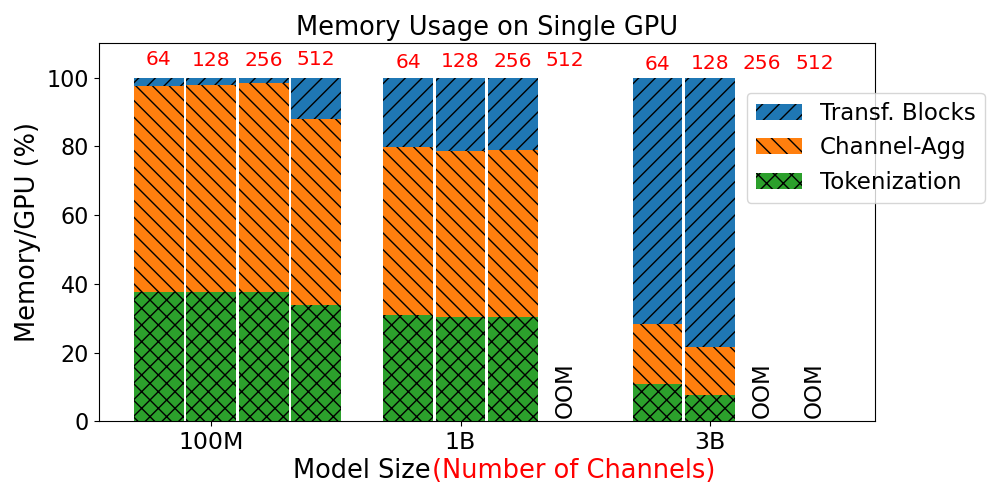}
\hspace*{0cm}\includegraphics[width=1.0\linewidth]{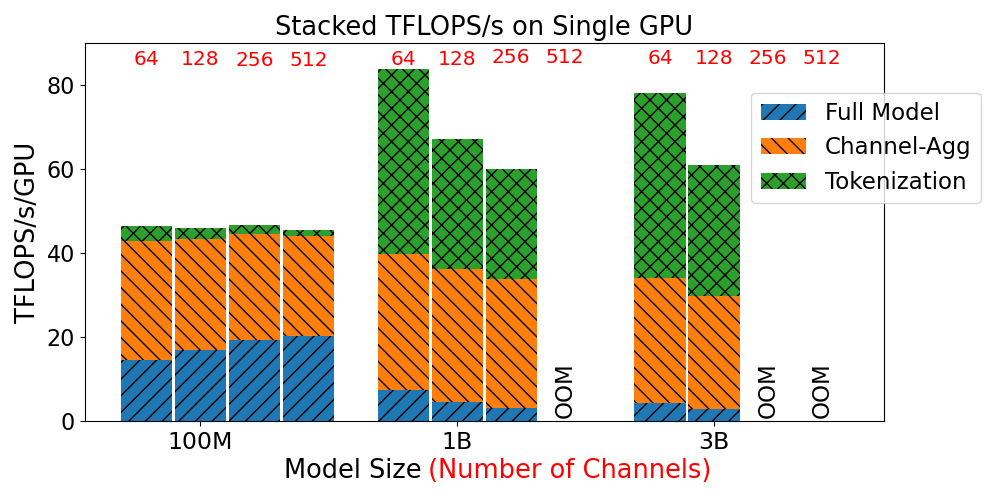}
\caption{The plots show memory usage and TFLOPs/GPU for single-GPU runs across the three main components of the model: tokenization, channel aggregation, and the transformer blocks. The top plot is normalized to the maximum memory usage of the full application. Three different model sizes were tested: 100M, 1B, and 3B. For the latter two, we observe out-of-memory (OOM) errors when using a large number of channels.}
\label{fig:perf-single-gpu}
\end{figure}

Figure~\ref{fig:perf-single-gpu} shows memory usage and compute performance for the three main components of the model: tokenization, channel aggregation, and the transformer blocks. The 100M-parameter model can handle up to 512 channels, while the 1B and 3B models can handle 256 and 128 channels, respectively. For the 100M and 1B parameter models, cross-attention and channel aggregation are the primary contributors to memory usage. In contrast, for the 3B parameter model, the transformer blocks dominate performance.

Although transformer blocks consume more memory as the model size increases, the majority of the compute (FLOPs) is directed toward channel aggregation and tokenization as the model grows. This highlights the need for an efficient strategy to distribute cross-channel compute across distributed tokens. As the plots show, for multi-channel datasets, tokenization and aggregation become key drivers of performance as the number of channels increases.

\subsection{Tensor Parallelism as Baseline} \label{tp-baseline}

Tensor parallelism (TP) distributes the embedding space of the model and is widely used to scale the model size of transformers. The TP approach used in this work is based on the method described in \cite{wang2024orbitoakridgebase}. Depending on the number of attention heads, individual layers are split and distributed across TP workers, with each GPU in the TP group computing only a portion of the layer parameters and activations. \emph{AllGather} operations are used to communicate parameters and activations between TP ranks, while the \emph{ReduceScatter} operation is used in the backward pass to allow each worker to compute the gradients for each layer.

In contrast, D-CHAG scales in the channel dimension, making the two methods complementary rather than directly comparable. However, we use TP as a baseline to demonstrate the performance gains achieved with our method, since D-CHAG is designed to work on top of model-parallel methods, but targets a different part of the FM architecture.

Furthermore, all of our performance experiments begin in the regime where TP is necessary, and model sharding techniques alone (such as FSDP) are insufficient for training a model. When FSDP alone is sufficient, it is generally preferable to scale in the batch dimension, as this reduces communication overhead and leads to better performance than any model-parallel approach. For example, we can use FSDP to train a 1.7B parameter model with up to 256 channels on two GPUs, or a 7B parameter model with 128 channels on a single node on Frontier.

\begin{figure}[h!]
\centering
\hspace*{-0.25cm}\includegraphics[width=0.99\linewidth]{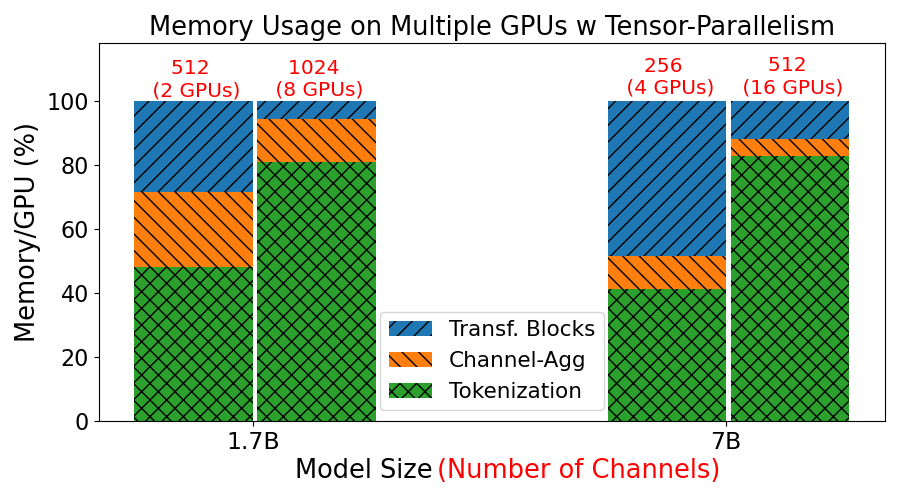}
\caption{The plots shows the memory usage per GPU of a 1.7B and a 7B parameter models using tensor-parallelism. The plots are normalized to the maximum memory usage of the full application.}
\label{fig:perf-tp}
\end{figure}

Figure~\ref{fig:perf-tp} shows the memory usage for 1.7B and 7B parameter models with respect to the channel dimensions that can only be trained using TP. Specifically, for the 1.7B parameter model, two GPUs are required to fit images with 512 input channels, while a full Frontier node is needed to fit images with 1024 channels using TP. On the other hand, for the 7B parameter model, images with 256 channels can fit on half of a Frontier node, while two Frontier nodes are required to fit images with 512 channels. For both models, we observe that tokenization and channel aggregation account from 50\% to 90\% of the memory usage when the number of channels is large.

These results are expected, as TP distributes the embedding dimension. As this dimension increases, the tokenization and channel aggregation layers also increase in size, but there is no existing implementation to distribute these parts of the model. It is important to note that these results are independent of the performance of the TP algorithm, as it only affects the transformer blocks. However, the absolute memory usage for tokenization and channel aggregation remains unchanged.

From these experiments, we conclude that, as in the single-GPU experiments, tokenization and channel aggregation are the most significant contributors to memory consumption, even when TP is used for images with a large number of channels.

\subsection{Distributed Tokenization Performance}

As mentioned in the previous section, one approach is to distribute the channel tokenization across TP ranks. This reduces the memory usage of tokenization, while also introducing a new \emph{AllGather} operation to communicate data in both the channel and spatial dimensions.

\begin{figure}[h!]
\centering
\hspace*{0cm}\includegraphics[width=1\linewidth]{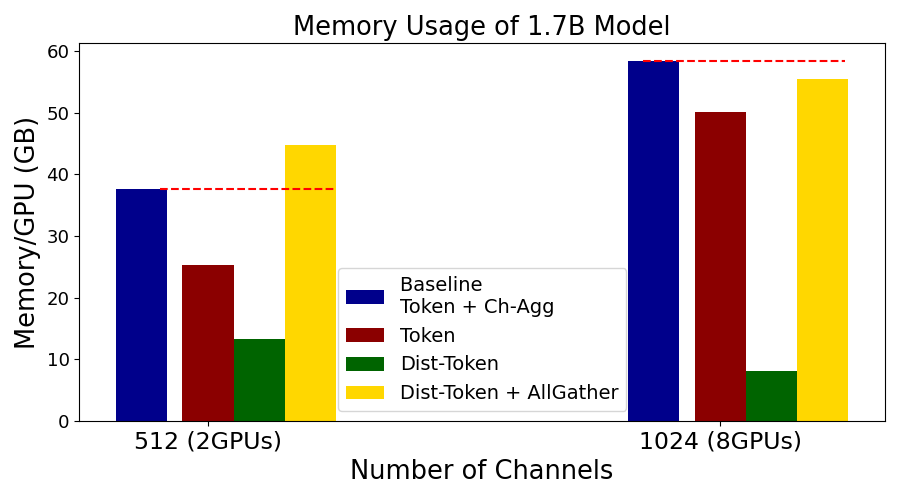}
\caption{The plots shows the memory usage per GPU for an 1.7B parameter model were the channel tokenization is distributed amonst TP ranks, as described in Section~\ref{dist-token-method}. The blue bars show the performance with tokenization plus channel aggregation, while the red bars represent the memory usage from tokenization alone. The green bars represent the performance from distributed tokenization, and the yellow bars show the performance from distributed tokenization and channel aggregation.}
\label{fig:perf-dist-token}
\end{figure}

As shown in Figure~\ref{fig:perf-dist-token}, distributing tokenization across TP ranks yields significantly better performance compared to tokenizing all channels within each TP rank (red bars vs. green bars). However, as discussed in Section~\ref{dist-token-method}, an \emph{AllGather} operation must be performed across both the channel and spatial dimensions to aggregate all channels. The yellow bars in Figure~\ref{fig:perf-dist-token} represent the memory usage for this part of the FM. When compared with the baseline TP performance (blue bars), the earlier benefits are effectively negated. In fact, for images with 512 channels, we observe a drop in performance, while for images with 1024 channels, only modest improvements are seen. This is primarily due to the \emph{AllGather} operation, which introduces substantial communication overhead across channel dimensions, causing the channel aggregation module to consume more memory than with TP alone.

\subsection{D-CHAG Performance}

Now that we have established that distributing tokenization alone does not lead to performance gains when applied to the full application, we proceed with experiments using the D-CHAG method, which combines distributed tokenization with hierarchical channel aggregation. The baseline for comparison is the TP method alone (as described in Section~\ref{tp-baseline}), versus D-CHAG combined with TP, using a fixed number of GPUs.

\begin{figure}[h!]
\centering
\hspace*{0cm}\includegraphics[width=1\linewidth]{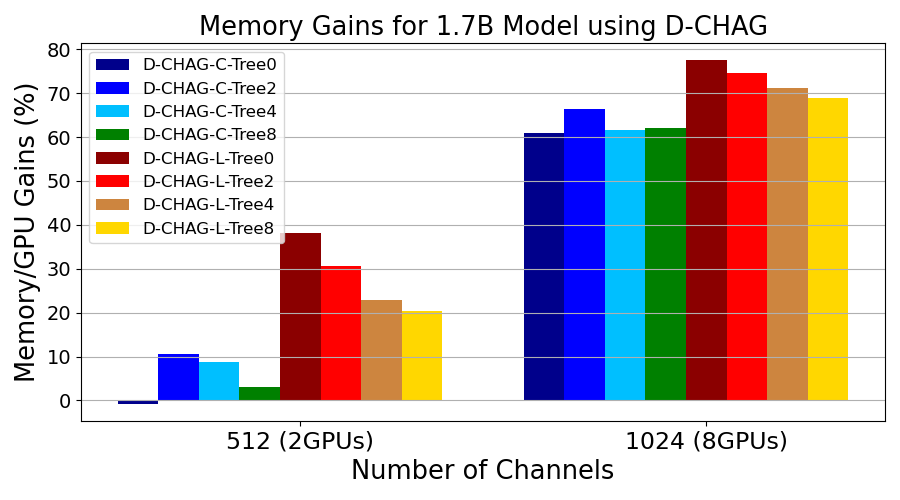}
\caption{The plots show, for a 1.7B-parameter model, the performance gains per GPU over the TP-only baseline across various configurations of the partial-channel aggregation module, as defined in Section~\ref{dchag-method}. \emph{Tree0} represents a single layer in the partial-channel aggregation module, while \emph{Tree2} represents two layers, and so on. The suffixes \emph{-C} and \emph{-L} indicate the type of layers used: in \emph{-C}, all layers are cross-attention, whereas in \emph{-L}, all layers are linear.}
\label{fig:perf-dchavit-2048}
\end{figure}

Figure~\ref{fig:perf-dchavit-2048} shows the measured performance of D-CHAG across various configurations of the partial-channel aggregation module, as defined in Section~\ref{dchag-method} and illustrated in Figure~\ref{fig:diagram-arch-dchavit}. For example, \emph{Tree2} for 512-channel images on two GPUs means that each GPU used two channel aggregation layers, with a maximum of 128 input channels per layer, while \emph{Tree8} used eight aggregation layers per GPU, with a maximum of 32 channels each. In all experiments presented in this work, the final channel aggregation layer—shared across TP ranks—is implemented as a cross-attention layer. The layers within the partial-channel aggregation module, however, were tested using both linear and cross-attention configurations.

As seen from the measured performance, using a single cross-attention layer results in slightly worse performance than the baseline for 512 channels, but yields a 60\% improvement for 1024 channels. As we deepen the hierarchical structure, we observe benefits even with 512-channel data, while the performance remains mostly constant for 1024-channel data.

On the other hand, when using linear layers, we see performance improvements even with a shallow hierarchical approach for both 512- and 1024-channel images. In fact, the best performance is achieved with \emph{D-CHAViT-L-Tree0}, which includes just one channel aggregation layer. Increasing the number of channel aggregation layers adds model parameters, introducing memory overhead. While additional layers appear beneficial for the 512-channel case, we find that using just one linear layer outperforms deeper configurations for both channel sizes.

For the remainder of the paper, we use the \emph{Tree0} configuration. We refer to the model as \emph{D-CHAG-L} when all layers in the partial aggregation module (see Fig.~\ref{fig:diagram-arch-dchavit}) are linear, and as \emph{D-CHAG-C} when cross-attention layers are used. Finally, in all experiments presented in this work, the final layer of the channel aggregation module has been implemented as a cross-attention layer.

\section{Evaluation}

In this section, we evaluate the D-CHAG method on two applications: weather forecasting and self-supervised mask prediction on plant hyperspectral images.

As mentioned earlier, although D-CHAG does not alter the core architecture of the FM model, incorporating the partial channel aggregation module results in a slightly larger number of parameters. The best-performing D-CHAG configuration uses linear layers to approximate the operations within the partial channel aggregation module, while retaining cross-attention for the final aggregation. The impact of these changes should be minimal, as they only introduce additional learnable parameters along the channel dimension. With appropriately chosen hyperparameters, we expect no degradation in performance. For simplicity, we tuned the hyperparameters for the baseline model only and kept them unchanged when applying the D-CHAG method, though further tuning may yield improved results.

For evaluation, we used single-GPU runs as the baseline. TP was used as our performance baseline, and as mentioned earlier, our results should be independent of the specific TP version, since TP and D-CHAViT are complementary. Therefore, we expect the measured performance gains to remain consistent across versions. When evaluating architectural changes, however, we focus on convergence behavior, which could be influenced by TP. To ensure a fair comparison, we treat single-GPU runs as a more reliable baseline. This approach allows us to simultaneously demonstrate the effects of the D-CHAG method and verify the overall correctness of the implementation, including the TP component.

\subsection{Mask Prediction on Hyperspectral Images}

\begin{figure}[h!]
\hspace*{0cm}\includegraphics[width=1.0\linewidth]{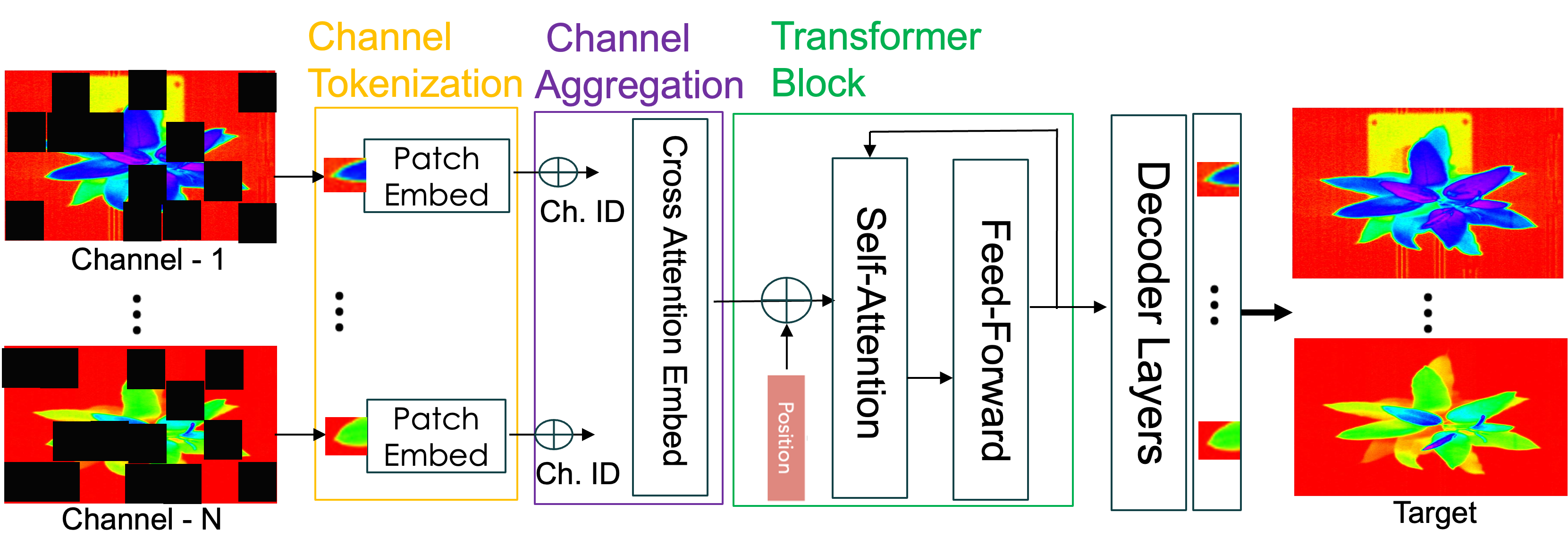}
\caption{A schematic of the full architecture used for self-supervised mask prediction on hyperspectral plant images.}
\label{fig:eval-arch-appl}
\end{figure}

Figure~\ref{fig:eval-arch-appl} shows the deep learning architecture used for self-supervised mask prediction of hyperspectral plant images. The model is based on the MAE architecture \cite{he2021maskedautoencodersscalablevision}, where tokens—representing image patches in this case—are masked, and the objective is to predict the missing content. This is a self-supervised approach in which the model learns the data distribution of the images.

\begin{figure}[h!]
  \centering
  \begin{tabular}{ c }
    \hspace*{-0.5cm}\includegraphics[width=.60\linewidth]{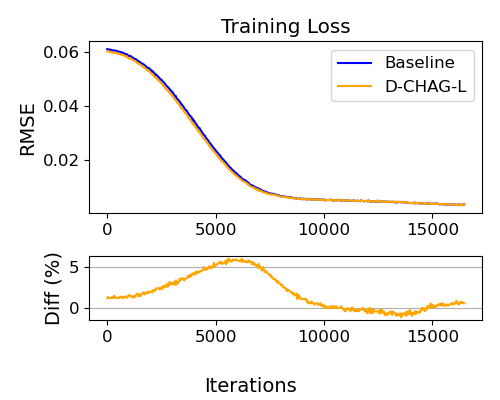}
  \end{tabular}%
  \begin{tabular}{ c c }
    \begin{subfigure}{0.15\textwidth}
        \includegraphics[width=\linewidth]{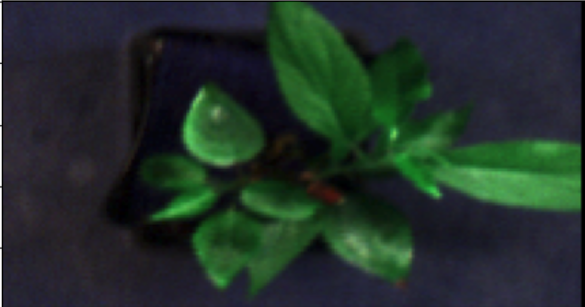}
        \caption{Original Image}
    \end{subfigure} \\
    \begin{subfigure}{0.15\textwidth}
      \includegraphics[width=\linewidth]{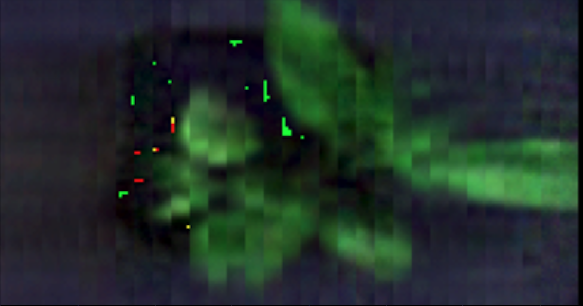}
      \caption{Mask Predicted}
  \end{subfigure}
  \end{tabular}
  \caption{The plot compares the training loss between the baseline and D-CHAG methods for the masked autoencoder application on hyperspectral plant images. Linear layers were used for the partial-channel aggregation module. For the baseline, training was performed on a single GPU, whereas the D-CHAG method was run on two GPUs. A 40M-parameter model was used with a batch size of 8. All hyperparameters were tuned for the baseline and kept identical for the D-CHAG runs.}
  \label{fig:convergence-appl}
\end{figure}

No pre-trained weights were used for training, and a dataset of visible-to-near-infrared (VNIR) hyperspectral images of Poplar—an important biomass feedstock for bioenergy research—was used. The dataset was collected by the Advanced Plant Phenotyping Laboratory (APPL) at Oak Ridge National Laboratory~\cite{APPL} and consists of 494 hyperspectral images, each with 500 spectral channels spanning wavelengths from 400 nm to 900 nm. This subset of 494 images was selected to maximize the variance explained by morphological and spectral plant phenotypes derived from RGB APPL imagery.

Figure~\ref{fig:convergence-appl} shows the results from the MAE model. Given the small size of the dataset, it is more appropriate to use iterations rather than epochs on the x-axis. For the partial-channel aggregation module, we used linear layers—referred to as \emph{D-CHAG-L}. As shown, there is good agreement in the training loss between the single-GPU implementation and the D-CHAG method (run on two GPUs) as training progresses. For this dataset, ground truth masks are not available. Therefore, in the right-hand plots of Figure~\ref{fig:convergence-appl}, we visualize the predicted full reconstructed image produced by the D-CHAG-trained model, compared to the original image, using pseudo-RGB representations of the hyperspectral data.

\subsection{Weather Forecasting}

Next, we examine a weather forecasting application. The FM architecture is similar to the previous one and is based on the ClimaX architecture \cite{nguyen2023climax}. However, instead of constructing the loss by masking the same images within each batch, the objective here is to predict a different image at a future timestep. The number of input channels is lower than in the case of hyperspectral images, but in this case, the 2D images are spatially uncorrelated, even though they correspond to the same timestep.

\begin{figure}[h!]
  \hspace*{-0.5cm}\includegraphics[width=1.05\linewidth]{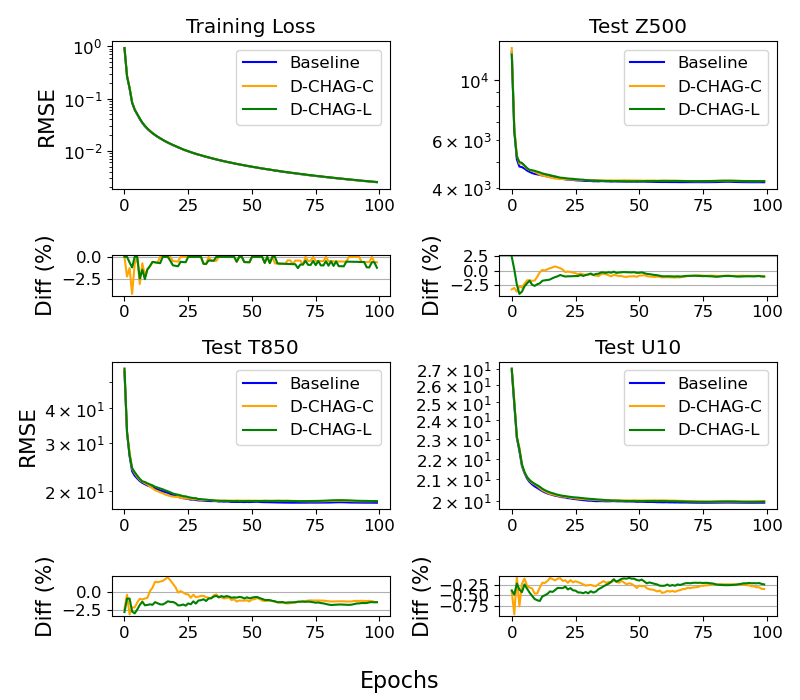}
\caption{The plot compares the training loss and three test RMSE variables between the baseline and the D-CHAG method for the weather forecasting application. Each epoch corresponds to a full pass through the ERA5 dataset. For the baseline, training was performed on a single GPU, whereas the D-CHAG method was run on four GPUs. A 53M-parameter model was used with a batch size of 512. All hyperparameters were tuned for the baseline and kept identical for the D-CHAG runs.}
\label{fig:convergence-era5}
\end{figure}

We used the well-established ERA5 weather dataset to test our method. For the ERA5 dataset, we selected five input atmospheric variables: geopotential, temperature, u-component of wind, v-component of wind, and specific humidity, each across more than 10 pressure levels. Additionally, we selected three surface-level variables: 2m temperature, 10m u-component of wind, and 10m v-component of wind. This combination results in a total of 80 variables (or channels). The performance is evaluated for three variables: geopotential at 500 hPa (Z500), temperature at 850 hPa (T850), and 10m u-component of wind (U10).

We also re-gridded the original dataset from ~0.25° (770 × 1440) to a 5.625° resolution (32 × 64). The re-gridding was performed using xESMF \cite{xesmf2020}, based on the high-performance Earth System Modeling Framework (ESMF), which supports re-gridding between general curvilinear grids with different algorithms, such as bilinear, nearest neighbor, and conservative. We applied the bilinear method for all our re-gridding tasks.

Figure~\ref{fig:convergence-era5} shows the weather forecasting results. We trained for up to 100 epochs on the full ERA5 dataset and tested on an independent one-year data set. Furthermore, we used four GPUs for the D-CHAG-C and D-CHAG-L models and one GPU for the baseline. As shown in the plot, the training loss matches almost exactly, while for the test RMSE, we observe only a 1\% lower rate.

\section{Performance at Scale}

In this section, we will measure the performance of the D-CHAG method as we scale the model size and the batch size. We will also present the best-performing configuration using a hybrid approach combining D-CHAG, TP, FSDP, and DP.

\subsection{Performance as Model-Size Scales}

As we mentioned before, the regime in which the D-CHAG method can show its advantage is where TP is necessary and model-sharding techniques alone aren't sufficient to fit the model into the memory of the GPU. For example, we can run a 7B parameter model with 128 channels on a single Frontier node using FSDP alone, but we can't fit 256 channels for the same model size. On a single Frontier node, we can only fit a 15B parameter model with up to 64 channels, while we can't fit a 26B parameter model on a single node at all.

\begin{figure}[h!]
\centering
\hspace*{0cm}\includegraphics[width=0.99\linewidth]{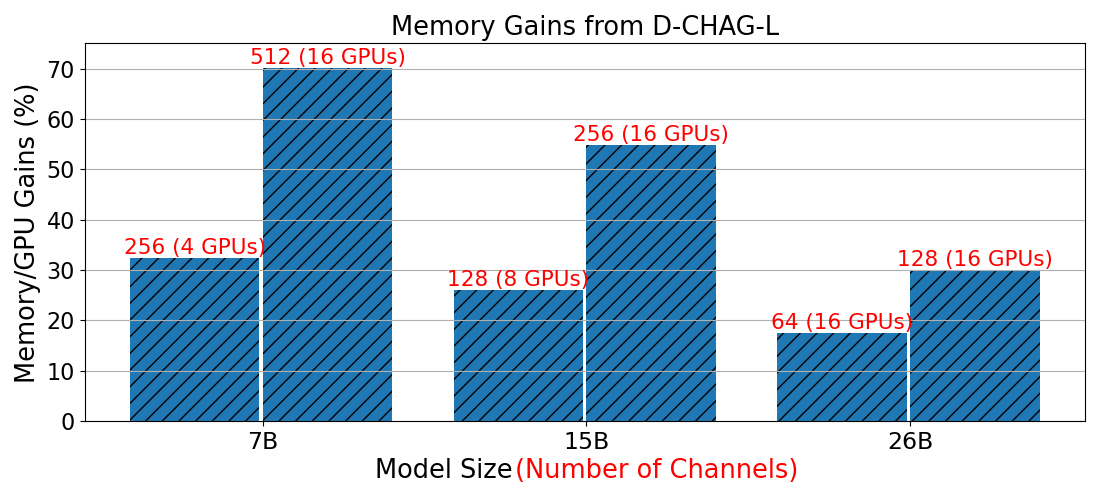}
\hspace*{0cm}\includegraphics[width=0.99\linewidth]{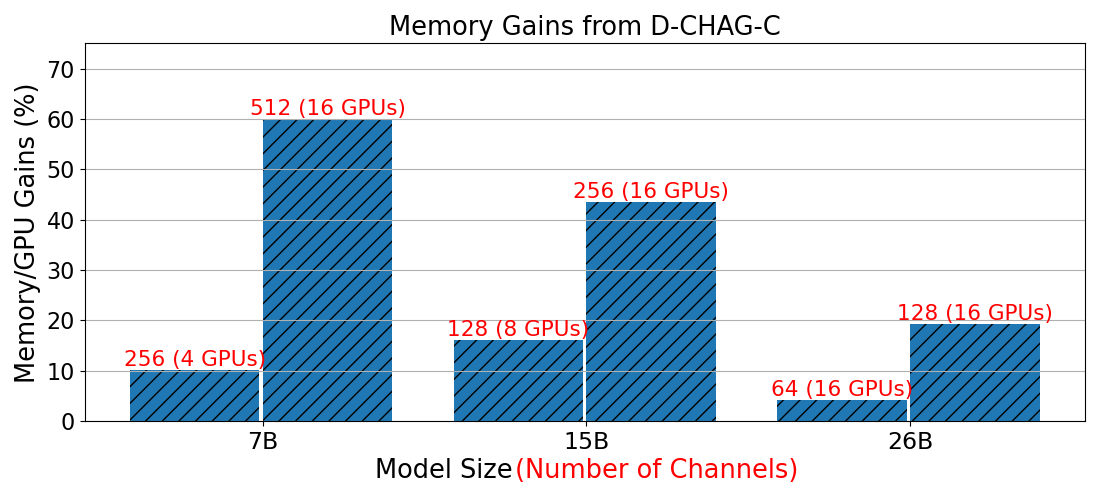}
\caption{The plots shows the performance gains per GPU for 7B, 15B and 26B parameter models when the D-CHAG method is combined with TP, versus TP alone.}
\label{fig:perf-dchavit-4096-and-up}
\end{figure}

Figure~\ref{fig:perf-dchavit-4096-and-up} shows the performance gains for three model sizes with channel configurations where TP is necessary. Specifically, it shows the performance gains of a 7B (4096 embedding, 32 layers, 32 attention heads), 15B (6144 embedding, 32 layers, 32 attention heads), and 26B (8192 embedding, 32 layers, 32 attention heads) model, using the D-CHAG method when combining it with TP, compared to TP alone.

We observe 30\% and 70\% improvements with the D-CHAG method when using linear layers in the partial-channel aggregation module, and 10\% and 60\% improvements when using cross-attention layers for a 7B parameter model. Furthermore, we measure performance gains of more than 20\% and 50\% for the 15B parameter model, and between 10\% and 30\% for the 26B parameter model.

We observe that using more linear layers instead of cross-attention layers results in better performance. This is expected, as the model size grows larger with more D-CHAG ranks, and linear layers have fewer parameters than cross-attention layers.

Additionally, for a fixed model size, we observe better performance gains as the number of channels increases. This is expected, as for the given architecture, when we increase the number of channels, the workload of the transformer block remains constant, while it increases for the tokenization and channel-aggregation modules. Moreover, as mentioned earlier, there is quadratic complexity between the number of channels and the channel aggregation module. Therefore, as the number of channels increases, the additional memory usage reduction from the cross-attention layers grows larger.

Finally, we can see that as the model parameters of the transformer blocks grow larger, the memory gains become smaller. This is a direct result of the previous point, where we see a larger gain as we increase the number of channels. As expected, with TP alone and for a fixed number of GPUs, as the model parameters increase, the number of channels that can be fitted becomes smaller. In fact, for the 26B parameter model, we were unable to fit a 256-channel image at all on Frontier.

\begin{figure}[h!]
\centering
\hspace*{0cm}\includegraphics[width=1.\linewidth]{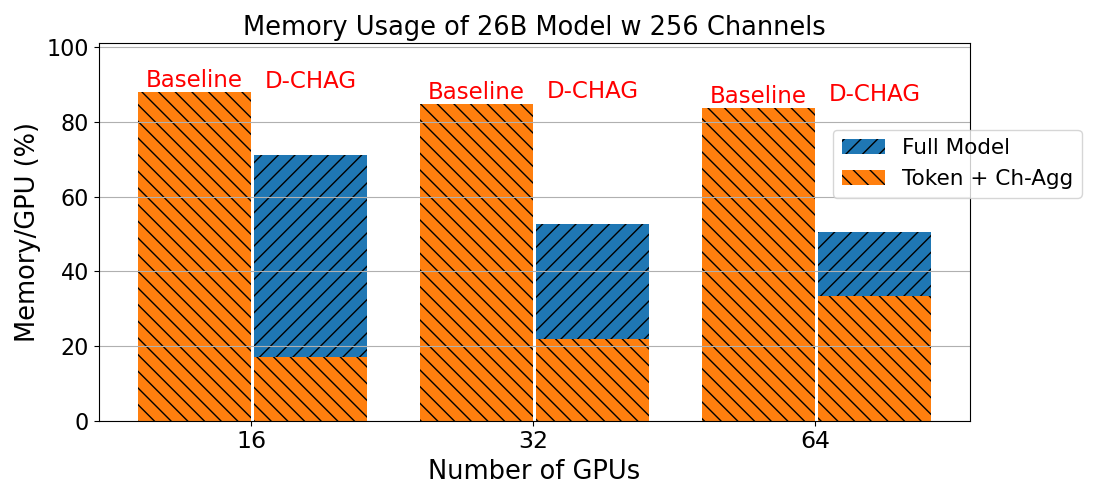}
\caption{The plots show the memory usage for a 26B parameter model with images containing 256 channels. The plots are normalized to the maximum memory capacity of Frontier's GPU. The baseline is the TP method alone, which isn't able to run the full model. For the D-CHAG and TP combined, the plot shows both the full model memory usage and the memory usage from tokenization and channel aggregation.}
\label{fig:perf-big}
\end{figure} 

Figure~\ref{fig:perf-big} shows that, at this scale, tokenization and channel-aggregation are prohibitively expensive for scaling a 26B parameter model with 256 channels using TP alone. As expected, using more GPUs won't help decrease memory usage because TP, or any other distributed method, doesn't address that. We can see a small decrease when using more GPUs because, as shown in Figure~\ref{fig:diagram-arch-tp}, TP distributes the embedding space of the channel aggregation module, but not in the channel dimension.

On the other hand, when using the D-CHAG method, we can fit a 26B parameter model with 512 channels, utilizing less than 80\% of the available memory. This demonstrates that our method enables the training of larger models for datasets with a high number of channels. As we scale beyond 26B, the constraints from tokenization and channel aggregation become more prominent, even with smaller numbers of channels, further showcasing the D-CHAG method’s ability to show its advantages to even fewer channels.

Finally, as shown in Figure~\ref{fig:perf-big}, D-CHAG has its own limitations as well. As we increase the number of GPUs, the self-attention (i.e., ViT part) is distributed more, and we observe larger benefits from TP. However, tokenization and channel aggregation increase. This is because, as we use more ranks, the layers from the D-CHAG method increase, leading to a larger model size. It is also worth noting that with our approach, the model size increases linearly, not quadratically as would have been the case in the spatial dimension.

\subsection{Performance Optimizations}

\begin{figure}[h!]
\centering
  \centering
  \hspace*{-.5cm}\includegraphics[width=1.\linewidth]{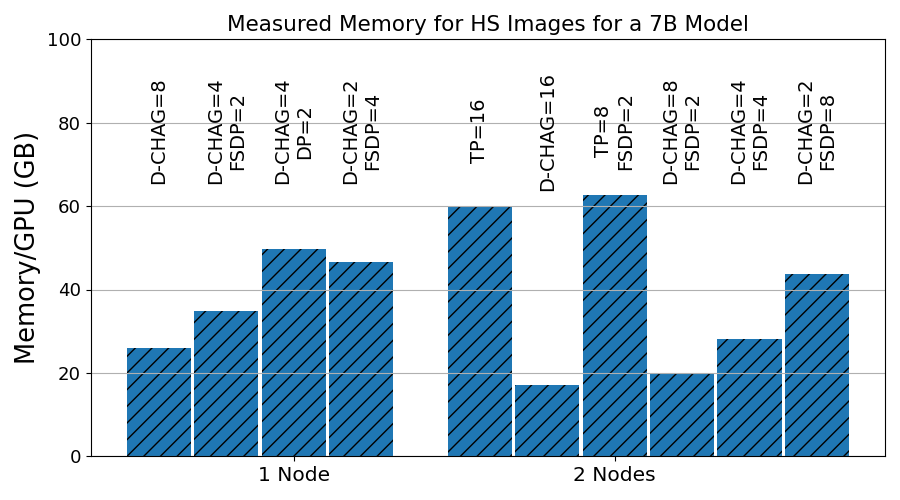}
  \centering
  \hspace*{-.5cm}\includegraphics[width=1.\linewidth]{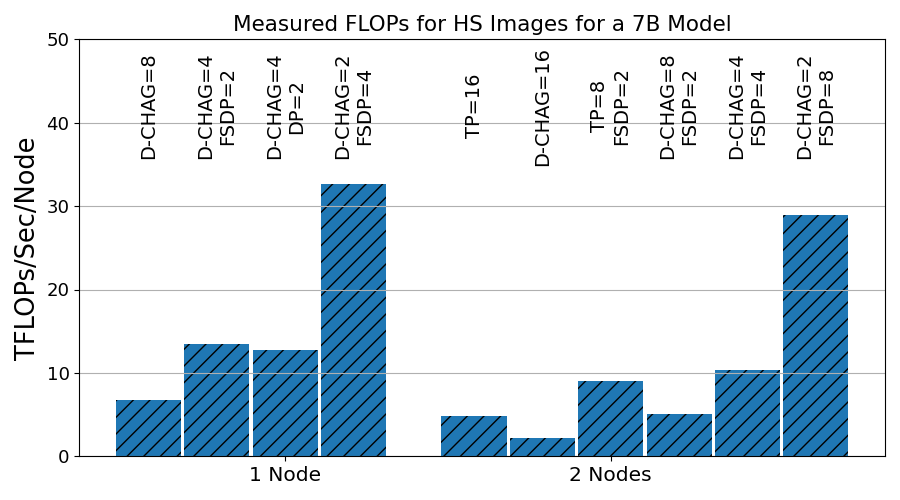}
\caption{The plots show the memory usage per GPU and the measured TFLOPs/sec/Node for different combinations between D-CHAG, TP, FSDP and DP for a 7B parameter model using images with 500 channels.}
\label{fig:perf-opt}
\end{figure}

Since we have already established the advantages and capabilities of the D-CHAG method, when combined with TP, we aim to find the optimal configuration by adding FSDP and DP for a fixed model size and compute budget. Our goal is to demonstrate better throughput using real hyperspectral images—the same ones used in the evaluation section—with 500 channels. We will choose a 7B parameter model and two Frontier nodes, which, as shown in Figure~\ref{fig:perf-dchavit-4096-and-up}, is the minimum number of GPUs required to fit 512 channels for this model size using TP alone.

Figure~\ref{fig:perf-opt} shows the memory usage per GPU and the measured TFLOPs/sec per Frontier node for images with 500 channels. With TP alone, we can only fit the 7B parameter model on two Frontier nodes. However, by using the D-CHAG method, we can fit the model on a single Frontier node, even with just two GPUs. As seen in the bottom plot of Figure~\ref{fig:perf-opt}, by reducing the model's memory requirements, we can achieve a higher number of TFLOPs/sec with the same number of resources by increasing the global batch size of the application.

\subsection{Performance as Batch-Size Scales}

\begin{figure}[h!]
\centering
  \centering
  \hspace*{-.5cm}\includegraphics[width=1.\linewidth]{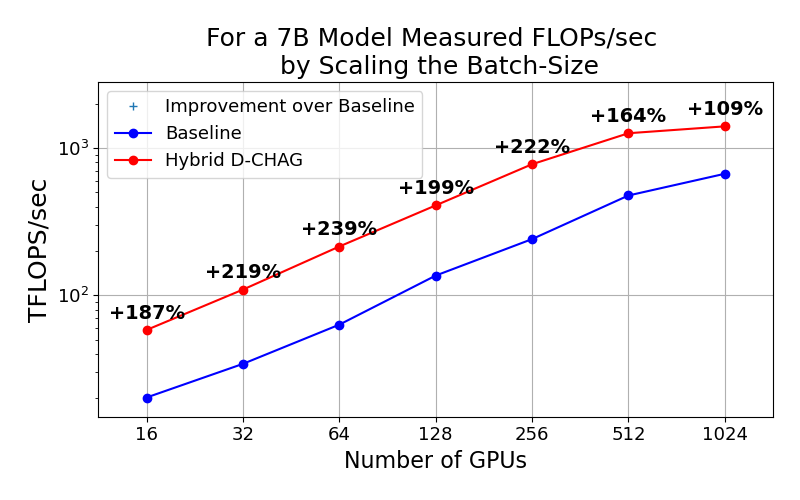}
\caption{Measured TFLOPs/sec for the full model for a fixed number of GPUs using the optimal configurations for both methods, as determined in Figure~\ref{fig:perf-opt}. A 7B parameter model was used for these runs, along with real hyperspectral images from APPL \cite{APPL}. For the baseline runs, TP, FSDP, and DP were used, while the D-CHAG method was combined with TP, FSDP, and DP (referred to as Hybrid D-CHAG). The percentage values in the red plot indicate the performance gains over the baseline.}
\label{fig:perf-scale}
\end{figure}

After determining the optimal configurations for both the Hybrid-D-CHAG method (i.e., D-CHAG, TP, and FSDP) and the baseline (i.e., TP and FSDP) using a 7B parameter model, we incorporated DP into both setups. This approach minimizes communication overhead as we scale the batch size. For the baseline, DP is applied in groups of two nodes, whereas the Hybrid-D-CHAG method allows for data parallelism across nodes.

Figure~\ref{fig:perf-scale} shows the total TFLOPs/sec achieved by the Hybrid-D-CHAG method compared to the baseline. As illustrated, the Hybrid-D-CHAG setup achieves more than double the sustained throughput when scaling batch size. Two main benefits contribute to increased FLOPs from the added memory efficiency: first, the ability to leverage data parallelism with fewer resources; and second, the earlier DP can be applied, the better scalability we achieve—since in DP, compute scales with communication.

Additionally, the Hybrid-D-CHAG approach pushes most of the heavy communication within the node, taking full advantage of the faster intra-node communication bandwidth. In contrast, the baseline must perform multiple \emph{AllGather} and \emph{ReduceScatter} operations across nodes, while D-CHAViT only requires a single \emph{AllReduce} across nodes at the end of the backward pass.

\section{Conclusion}

In this paper, we propose a novel distributed approach for training foundation models on multi-channel datasets that significantly reduces memory usage and enhances computational performance. The D-CHAG method is independent of the ViT architecture or the model-parallel strategy used and enables training of larger models on datasets with a high number of channels by efficiently distributing the tokenization and channel aggregation layers.
For datasets with a large number of channels that can already leverage existing model-parallel methods, we observe up to 75\% memory usage reduction when combined with D-CHAG.

    
\section*{Acknowledgments}
This manuscript has been authored by UT-Battelle, LLC, under contract DE-AC05-00OR22725 with the US Department of Energy (DOE). The US government retains and the publisher, by accepting the article for publication, acknowledges that the US government retains a nonexclusive, paid-up, irrevocable, worldwide license to publish or reproduce the published form of this manuscript, or allow others to do so, for US government purposes. DOE will provide public access to these results of federally sponsored research in accordance with the DOE Public Access Plan (http://energy.gov/downloads/doe-public-access-plan). An award of computer time was provided by the INCITE program. This research used resources of the Oak Ridge Leadership Computing Facility, which is a DOE Office of Science User Facility supported under Contract DE-AC05-00OR22725. Use of the Advanced Plant Phenotyping Laboratory is acknowledged at Oak Ridge National Laboratory. This work was funded by the Center for Bioenergy Innovation (CBI), which is a U.S. Department of Energy Bioenergy Research Center supported by the Office of Biological and Environmental Research in the DOE Office of Science. Oak Ridge National Laboratory is managed by UT-Battelle, LLC for the US DOE under Contract Number DE-AC05-00OR22725.



\bibliographystyle{ACM-Reference-Format}
\bibliography{sample-base}

\end{document}